\documentclass{article}

\usepackage[preprint]{neurips_2025}
\usepackage{caption}

\usepackage[utf8]{inputenc} 
\usepackage[T1]{fontenc}    
\usepackage{hyperref}       
\usepackage{url}            
\usepackage{booktabs}       
\usepackage{amsfonts}       
\usepackage{nicefrac}       
\usepackage{microtype}      
\usepackage{xcolor}         
\usepackage{graphicx}
\usepackage{enumitem}
\usepackage{listings}
\usepackage[T1]{fontenc}
\usepackage[scaled=.8]{beramono}   
\usepackage{soul}
\usepackage{xspace}
\usepackage{amsmath}
\usepackage{tablefootnote}
\sethlcolor{white}

\usepackage{algorithm}
\usepackage{algpseudocode}
\usepackage{amsmath}

\newcommand\mypar[1]{\par\noindent\textbf{#1}\;\;}
\newcommand{\METHODNAME}{{\fontfamily{txtt}\selectfont {RIPT-VLA}}\xspace}
\newcommand{\METHODNAMESHORT}{{\fontfamily{txtt}\selectfont {RIPT}}\xspace}

\algrenewcommand{\algorithmicrequire}{\textbf{Input:}}
\let\Input\Require

\newcommand{\eg}{\textit{e.g.}\xspace}

\usepackage{xcolor}
\definecolor{myred}{RGB}{200,66,25}
\title{Interactive Post-Training for \\ Vision-Language-Action Models}

\definecolor{betterblue}{RGB}{30, 100, 200}

\author{Shuhan Tan$^1$, Kairan Dou$^2$, Yue Zhao$^1$, Philipp Krähenbühl$^1$\\
UT Austin$^1$, Nankai University$^2$\\
\texttt{\href{https://ariostgx.github.io/ript_vla/}{Code \& Model: \textcolor{betterblue}{https://ariostgx.github.io/ript\_vla/}}}
}

\begin{document}

\maketitle

\begin{abstract}
We introduce \METHODNAME, a simple and scalable reinforcement-learning-based interactive post-training paradigm that fine-tunes pretrained Vision-Language-Action (VLA) models using only sparse binary success rewards.
Existing VLA training pipelines rely heavily on offline expert demonstration data and supervised imitation, limiting their ability to adapt to new tasks and environments under low-data regimes.
\METHODNAME addresses this by enabling interactive post-training with a stable policy optimization algorithm based on dynamic rollout sampling and leave-on-out advantage estimation.
\METHODNAME has the following characteristics.
First, \METHODNAME applies to various VLA models, resulting in an improvement on the lightweight QueST model by 21.2\%, and the 7B OpenVLA-OFT model to an unprecedented \textbf{97.5\%} success rate.
Second, \METHODNAME is computationally efficient and data-efficient: With only one demonstration, \METHODNAME enables an unworkable SFT model (\textbf{4\%}) to succeed with a \textbf{97\%} success rate within \textbf{15} iterations.
Furthermore, we demonstrate that the policy learned by \METHODNAME generalizes across different tasks and scenarios and is robust to the initial state context.
These results highlight \METHODNAME as a practical and effective paradigm for post-training VLA models through minimal supervision.
\end{abstract}

\vspace{-1em}

\begin{figure}[ht]
    \centering
    \includegraphics[width=0.95\textwidth]{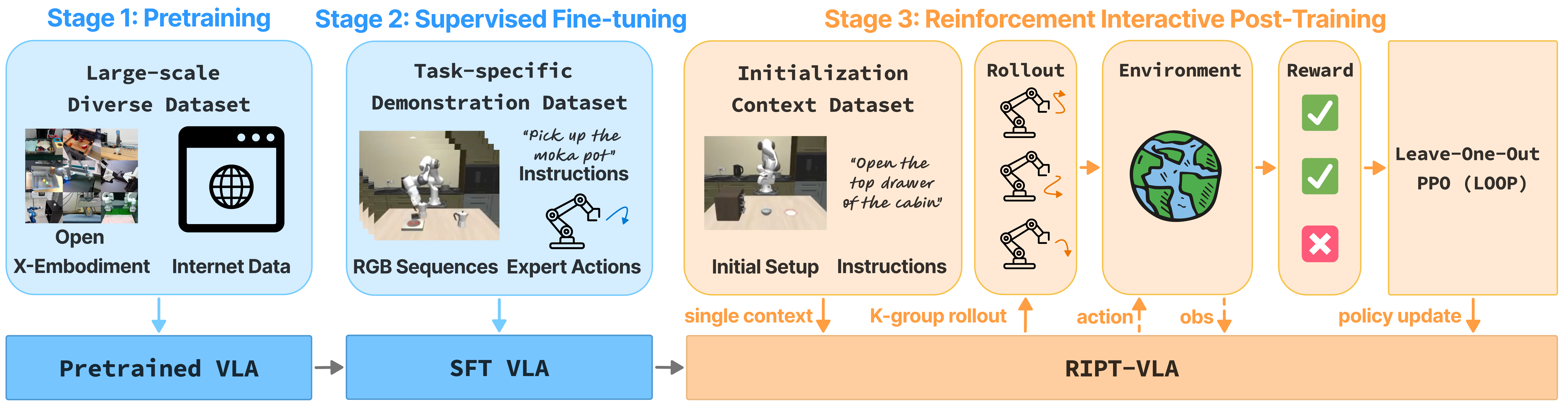} \\
    \includegraphics[width=0.19\textwidth]{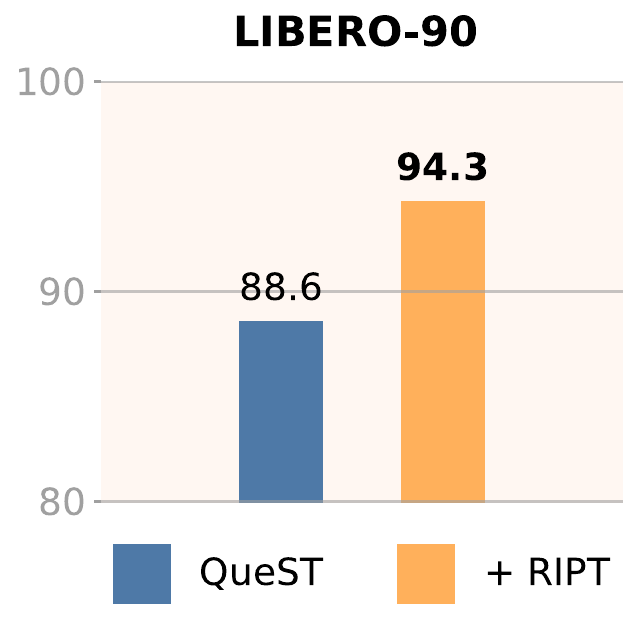}
    \includegraphics[width=0.19\textwidth]{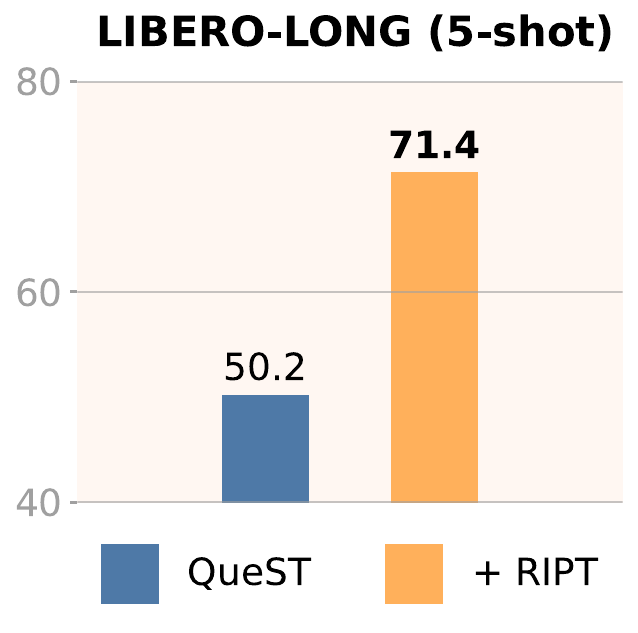}
    \includegraphics[width=0.19\textwidth]{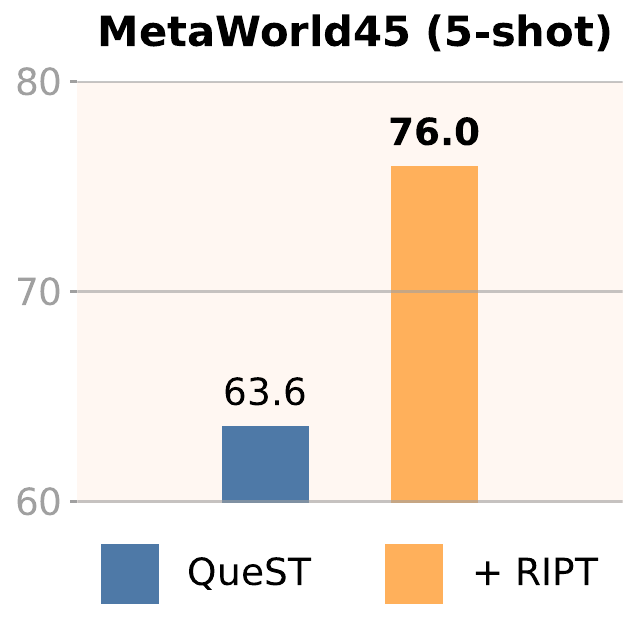}
    \includegraphics[width=0.19\textwidth]{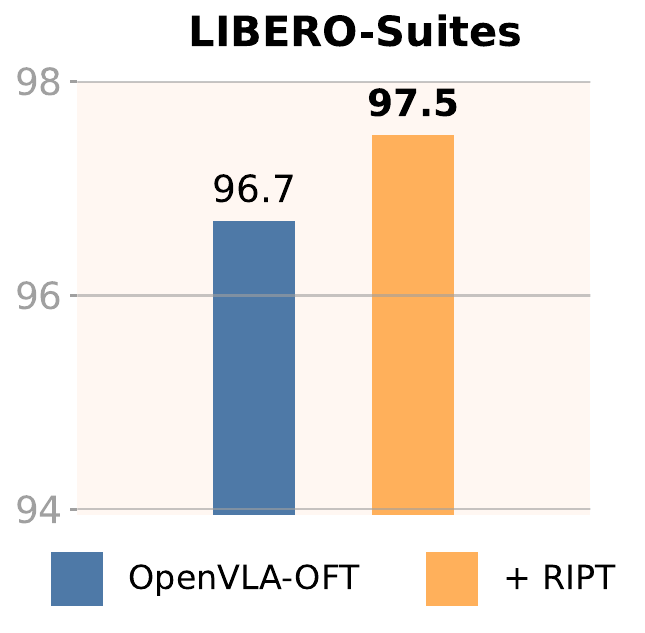}
    \includegraphics[width=0.19\textwidth]{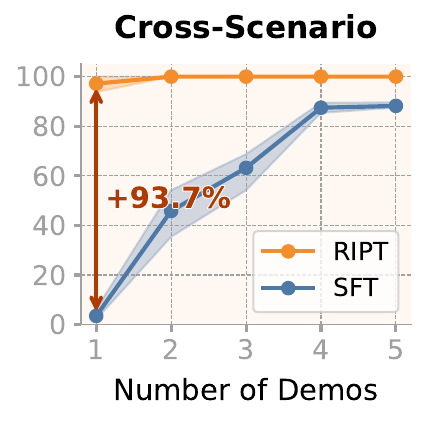}
    \caption{Overview of \METHODNAME. 
    While VLA models are typically trained with two supervised stages, we propose a third stage: Reinforcement Interactive Post-Training for VLA.
    \METHODNAME sets state-of-the-art results across diverse benchmarks. It also presents remarkable improvement under the low-data regime, transforming an SFT model from near failure to 97\% with one demonstration. 
    }
    \label{fig:main_figure}
\end{figure}

\section{Introduction}
Vision-Language-Action (VLA) models~\cite{pmlr-v229-zitkovich23a} aim to enable agents to perceive, reason, and act in the physical world with a unified interface.
Current VLA models are trained with two supervised stages: 
large-scale pretraining on diverse human demonstrations, followed by supervised fine-tuning (SFT) on smaller-scale task-specific datasets.
This paradigm has some distinct advantages: 
Pre-training enables the VLA model to build general visuomotor skills while SFT allows it to specialize in specific environments~\cite{kim2024openvla}.
Supervised training allows VLA models to learn from large-scale pre-recorded vision-language-action datasets.
However, this supervised approach also has two core limitations:
First, data are collected offline. The VLA model learns to imitate interactions with the environment, but never sees the consequences of its own actions.
As a result, the learned policy often fails to handle the complexities of real-world scenarios, especially for long-horizon tasks.
Second, task-specific SFT via imitation learning relies heavily on large-scale, high-quality human demonstrations.
These data are expensive and time-consuming to collect, and performance degrades significantly when only a small number of demonstrations are available. 

In this paper, we propose \textbf{\METHODNAME}: a third stage for VLA training paradigm with \textbf{R}einforcement \textbf{I}nteractive \textbf{P}ost-\textbf{T}raining.
After pretraining and supervised fine-tuning, we allow the VLA model to interact with the multitask environment and receive binary success/failure rewards.
We then optimize the VLA model to directly improve its success rate across multiple tasks through reinforcement learning.
Inspired by prior RL frameworks for LLMs reasoning~\cite{guo2025deepseek}, we propose a stable and efficient RL framework for VLA finetuning in Section~\ref{sec:method}.
Specifically, we extend the LOOP framework~\cite{chen2025reinforcementlearninglonghorizoninteractive} which combines REINFORCE leave-one-out (RLOO) advantage estimation~\cite{koolattention} and proximal policy optimization (PPO)~\cite{schulman2017proximal}.
Unlike LOOP, we construct uniform batches of non-zero advantage samples, filtering out any group of trajectories with zero-advantage, and sampling rollouts until sufficient samples exist.
This uniform batch construction leads to improved training stability, especially as training progresses and the VLA becomes more successful.
\textbf{\METHODNAME} allows efficient and stable VLA policy update \textit{without} relying on shaped or learned rewards, or critic models.
Using Reinforcement Learning in a third training stage has a few distinct advantages:
It is more data efficient, yielding close to state-of-the-art performance with only a single SFT demonstration.
The resulting VLA model achieves a much higher performance on the end-task, as it gets to see interactions with the environment during training.
\METHODNAME works with both tokenized~\cite{mete2024quest} and continuous actions~\cite{kim2025openvlaoft}.

\METHODNAME resonates with the recent trend of paradigm shift in LLM training~\cite{ouyang2022training,guo2025deepseek}.
While pretraining on large-scale text corpora equips LLMs with broad knowledge and powerful skills, they often struggle with challenging tasks that require precise reasoning, multi-step planning, or tool using~\cite{wang2024q}. 
To address these limitations, reinforcement learning has emerged as a critical post-training stage to reactivate and steer pretrained knowledge with only a small amount of interactive feedback~\cite{ouyang2022training}.
Similarly, we observe that pretrained VLA models also encode rich visuomotor skills, yet struggle to apply them effectively for new tasks and scenarios.
\METHODNAME bridges this gap by using only sparse binary rewards to unlock these latent skills with a small number of optimization steps.

Through comprehensive experiments in Section~\ref{sec:exp}, we demonstrate that \METHODNAME achieves state-of-the-art results when combined with both large-scale and lightweight VLA models across a diverse set of tasks.
On the LIBERO benchmark~\cite{liu2023libero}, \METHODNAME improves QueST~\cite{mete2024quest}, the best lightweight VLA model, on all four task suites by {\bf 10.9\%} absolute success rate (SR) on average (Table~\ref{tab:libero-four-tasks}).
When evaluated on OpenVLA-OFT~\cite{kim2025openvlaoft}, the best-performing large VLA model with an already high success rate (96.7\%), \METHODNAME still helps by further reducing the failure rate from 3.3\% to 2.5\%.
We also achieve top performance on many-task benchmarks LIBERO-90 \textbf{(94.3\%)} and MetaWorld45~\cite{yu2020meta} (\textbf{92.2\%}), showing the effectiveness of \METHODNAME in improving multi-task (up to 90) performance with a single model (Table~\ref{tab:all-benchmark-results}).
Most notably, in the extreme low-data regime with only a single training demo, \METHODNAME adapts pretrained knowledge to new task goals or scenarios with remarkable efficiency, boosting success rate from below 4\% to over 97\% within only 15 RL iterations.

\section{Related Works}

\mypar{Vision-Language-Action Models.}
Vision-Language-Action (VLA) models empower embodied agents to interpret multimodal inputs—such as visual observations and natural-language instructions—and translate them into meaningful actions within the physical world~\cite{pmlr-v229-zitkovich23a}.  
Seminal works like RT-2~\cite{pmlr-v229-zitkovich23a}, RT-1~\cite{brohan2022rt}, PaLM-E~\cite{driess2023palm}, Octo~\cite{team2024octo}, Dita~\cite{hou2025dita}, $\pi_0$~\cite{black2024pi0}, and $\pi_{0.5}$~\cite{intelligence2025pi_}, together with OpenVLA~\cite{kim2024openvla}, showcase VLAs achieving emergent semantic reasoning and generalization to novel tasks and environments.  
These models are typically developed through a two-stage supervised-learning paradigm that begins with an initial pre-training phase on extensive, web-scale datasets~\cite{driess2023palm, rt12022arxiv}, which is crucial for acquiring generalizable visuomotor skills, grounding language in perception, and building robust internal representations.  
While this two-stage approach has advanced the field, its offline nature imposes key limitations.  
The supervised fine-tuning (SFT) stage typically requires vast expert demonstrations for new tasks or environments, thereby degrading few-shot performance.  
This highlights a critical gap: the need for methods that adapt pretrained VLAs beyond static imitation by leveraging interactive experience and reducing reliance on extensive expert data.

\mypar{Reinforcement Learning for Large Language Models.}
Large Language Models (LLMs) offer a precedent for enhancing pretrained models.  
While LLMs gain broad capabilities via pre-training and SFT, they often struggle with complex reasoning, planning, or constraint satisfaction~\cite{wang2024q}.  
To address this, Reinforcement Learning (RL) has emerged as a transformative third stage in LLM training—enabling learning from interactive feedback rather than static datasets~\cite{ouyang2022training}.  
Recent progress shows RL can unlock latent capabilities for math~\cite{lightman2023let, shao2024deepseekmath}, self-verifiable proofs~\cite{liu2024direct}, long-horizon planning through tree-of-thoughts~\cite{yao2023tree}, and preference-aligned generation with AI feedback~\cite{lee2023rlaif}.  
This paradigm, in which pretrained knowledge is steered by targeted feedback, strongly motivates a similar approach for Vision-Language-Action models: RL has the potential to adapt pretrained VLAs more effectively to the interactive and consequential nature of embodied tasks.

\mypar{Reinforcement Learning for VLA.}
Recent works have explored applying reinforcement learning to pretrained VLA models to overcome limitations of supervised fine-tuning and adapt to novel tasks without collecting new demonstrations.
iRe-VLA~\cite{liu2024irevla} addresses optimization instability by alternating between PPO-based updates on a frozen VLM backbone and supervised distillation stages.
However, it still relies on a learned value critic during PPO, and requires shaped reward functions or success weighting to guide policy learning.
ConRFT~\cite{chen2025conrft} further combines offline Q-learning with online consistency-policy updates, but similarly depends on a parameterized value function.
Both methods require careful coordination between offline and online stages to stabilize critic learning.
In contrast, \METHODNAME introduces a fully critic-free optimization framework with simpler training dynamics under sparse binary rewards.

\section{Preliminary}
\subsection{Vision-Language-Action Models}
\label{sec_pre_vla}
\mypar{Autoregressive VLA rollout.} 
A vision-language-action (VLA) model $\pi_\theta$ maps a sequence of observations and previous actions $(o_{1:t}, a_{1:t-1})$, along with a natural language goal $g$, to a probability distribution over the next action $a_t$. These models operate autoregressively: $a_t \sim \pi_\theta(\cdot \mid o_{1:t}, g, a_{1:t-1})$.
Given an initial observation-goal pair context $\mathbf{c} = (o_1, g)$, the model generates a sequence of actions conditioned on past information in an autoregressive way:
\begin{equation}
\pi_\theta(a_{1:T} \mid o_{1:T}, g) = \prod_{t=1}^T \pi_\theta(a_t \mid o_{1:t}, g, a_{1:t-1}).
\end{equation}
We denote this sampling process as $\mathbf{a} = a_{1:T} \sim \pi_\theta(\cdot \mid \mathbf{c})$, the observation sequence as $\mathbf{o} = o_{1:T}$.
Sequences terminate upon task success or reaching a time limit.
For each rollout sequence and task goal $g$, the environment $\mathcal{E}$ returns a binary reward $R=1$ when the task goal is successfully reached, and $R=0$ otherwise.
The environment $\mathcal{E}$ can be either a simulator~\cite{yu2020meta,liu2023libero} or the real world.

There are two common ways of action prediction in VLA models. 
The \textit{tokenized action head} represents actions as discrete tokens from a fixed vocabulary and predicts actions via classification over the token set.
In contrast, the \textit{regression action head} directly predicts real-values action vectors via regression.

\mypar{Current VLA training paradigm.}
Current Vision-Language-Action (VLA) models are typically trained in two stages: \textbf{Stage 1: Pretraining} and \textbf{Stage 2: Supervised Fine-tuning}.

In Stage 1, a base policy $\pi_\theta$ is pretrained on a large-scale, diverse dataset of real-world demonstrations, denoted by $\mathcal{D}_{\text{pretrain}} = \{(\mathbf{o}, \mathbf{a}, g)\}_{i=1}^N$.
The policy is trained to imitate the ground-truth actions given offline data in $\mathcal{D}_{\text{pretrain}}$.
For VLA models with a tokenized action head, the loss is:
\begin{equation}
\mathcal{L}_{\text{pre}}(\theta) = -\mathbb{E}_{(\mathbf{o}, \mathbf{a}, g) \sim \mathcal{D}_{\text{pretrain}}} \left[ \sum_{t=1}^T \log \pi_\theta(a_t \mid o_{1:t}, g, a_{1:t-1}) \right],
\label{equ:sft}
\end{equation}
while for regression action head $\mathcal{L}_{\text{pre}}(\theta)$ is implemented as an MSE or L1 loss. 
This stage enables VLA models to capture strong representations and learn general visuomotor and instruction-following capabilities.

In Stage 2, the pretrained policy is supervised fine-tuned on a smaller, multitask dataset to improve performance on a small set of target tasks, denoted by $\mathcal{D}_{\text{sft}} = \{(\mathbf{o}, \mathbf{a}, g)\}_{i=1}^{N'}$.
Typically, $\mathcal{D}_{\text{sft}}$ contains around 50 high-quality human demonstrations per task~\cite{mete2024quest}.
The VLA model is trained with the same objective function as in Stage 1.
This stage enables the model to adapt its learned skills from Stage 1 to a specialized set of skills for the target tasks.

Although being the standard process of VLA training, this two-stage process has two significant issues. Firstly, it relies only on offline supervision and lacks interactive feedback from the environment. Therefore, the learned policy may often fail in real rollouts due to distribution shift and cascading errors, especially for long-term rollouts.
Furthermore, the performance of VLA heavily relies on the high quality and quantity of the task-specific data in $\mathcal{D}_{\text{sft}}$, which is often hard and costly to obtain.

\mypar{VLA as Markov decision processes.}
To better optimize VLA models, we define its task as a Markov decision process (MDP). 
Each episode is initialized with a context $\mathbf{c} = (o_1, g)$.
The \textit{state} is represented as $[o_{1:t}, g, a_{1:t-1}]$, which includes the language goal $g$, the sequence of past observations $o_{1:t}$, and past actions $a_{1:t-1}$. 
At each timestep $t$, the VLA policy produces an \textit{action} sampled from the policy distribution: $a_t \sim \pi_\theta(\cdot \mid o_{1:t}, g, a_{1:t-1})$.
The environment transitions to the next observation $o_{t+1}$ based on hidden environment dynamics, producing a new state $[o_{1:t+1}, g, a_{1:t}]$.
After a sequence of actions $a_{1:T}$, the agent receives a binary \textit{reward} $R(\mathbf{c}, \, \mathbf{a}) \in \{0, 1\}$ from the environment $\mathcal{E}$, indicating task success or failure. 
The objective of VLA optimization is essentially learning a policy $\pi_\theta$ that maximizes expected task success reward:
\begin{equation}
R_\theta(\mathbf{c}) = \mathbb{E}_{\mathbf{a} \sim \pi_\theta(\cdot \mid \mathbf{c})} \left[ R(\mathbf{c}, \mathbf{a}) \right].
\label{equ:reward_obj}
\end{equation}

\subsection{Reinforcement Policy Optimization}
\label{sec_pre_loop}
We consider the reinforcement learning setting where an agent interacts with an environment $\mathcal{E}$ to learn a policy $\pi_\theta(\mathbf{a} \mid \mathbf{c})$ that maximizes the expected return:
$\mathbb{E}_{\mathbf{c} \sim \mathcal{D}_{\text{context}},\, \mathbf{a} \sim \pi_\theta}[R(\mathbf{c}, \mathbf{a})],$
where $\mathbf{c}$ is the context (e.g., goal and initial observation), $\mathbf{a}$ is a trajectory (e.g., sequence of actions), and $R(\mathbf{c}, \mathbf{a}) \in \{0,1\}$ is a sparse binary reward returned by the environment.
To optimize this objective, a standard approach is policy gradient, which updates $\pi_\theta$ with:
\begin{equation}
\nabla_\theta L_\theta(\mathbf{c}) = \mathbb{E}_{\mathbf{a} \sim \pi_\theta}[\nabla_\theta \log \pi_\theta(\mathbf{a} \mid \mathbf{c}) \cdot A(\mathbf{c}, \mathbf{a})],
\end{equation}
where $A(\mathbf{c}, \mathbf{a})$ is the advantage function indicating how much better the action $\mathbf{a}$ is compared to a baseline.
In practice, it is hard to compute $A(\mathbf{c}, \mathbf{a})$, especially under sparse rewards. 
To address this issue, a recent work proposed a critic-free optimization framework called Leave-One-Out Proximal Policy Optimization (\textbf{LOOP})~\cite{chen2025reinforcementlearninglonghorizoninteractive}.
Specifically, it combines the two methods below.

\mypar{Leave-One-Out Advantage Estimation (RLOO)~\cite{koolattention}.}
For each sampled context $\mathbf{c}$, we draw $K$ rollouts $\{\mathbf{a}_k \sim \pi_\psi(\cdot \mid \mathbf{c})\}_{k=1}^K$ under a fixed sampling policy $\pi_\psi$. Each rollout receives a binary reward $R_k = R(\mathbf{c}, \mathbf{a}_k)$. The leave-one-out baseline for rollout $k$ is computed by averaging the other rewards:
\begin{equation}
b_k = \frac{1}{K - 1} \sum_{j \ne k} R_j,
\quad A_k = R_k - b_k.
\label{equ_rloo}
\end{equation}
This group-normalized advantage indicates how much better or worse a rollout performance is relative to others given the same context.
This allows us to efficiently compute a stable advantage signal from sparse binary rewards, without requiring learning value functions.

\mypar{Proximal Policy Optimization (PPO)~\cite{schulman2017proximal}.}
To update $\pi_\theta$ using collected rollouts $\{(\mathbf{c}_k, \mathbf{a}_k, A_k)\}$, we compute the importance ratio $r_k = \pi_\theta(\mathbf{a}_k \mid \mathbf{c}_k)/\pi_\psi(\mathbf{a}_k \mid \mathbf{c}_k)$, where $\pi_\theta$ is the current updating policy and $\pi_\psi$ is the fixed sampling policy (normally set to the latest checkpoint of $\pi_\theta$). We then optimize $\pi_\theta$ with the following clipped objective:
\begin{equation}
\mathcal{L}_{\text{PPO}} = -\min \left( r_i A_i,\ \text{clip}(r_i,\ 1 - \epsilon,\ 1 + \epsilon) A_i \right),    
\label{equ_ppo}
\end{equation}
where $\epsilon$ is a small updating threshold (we use 0.2). 
This objective encourages rollouts with positive advantages while preventing unstable updates when $\pi_\theta$ deviates too far from its previous version $\pi_\psi$.

LOOP adopts PPO to optimize the advantage estimated by RLOO, which enables sample-efficient policy optimization in sparse reward settings without critics.
It serves as an out-of-box working implementation for our interactive post-training framework in Section~\ref{sec:method}.

\section{\METHODNAME}
\label{sec:method}

\begin{algorithm}[t]
\caption{\textbf{\METHODNAME}: \textbf{R}einforcement \textbf{I}nteractive \textbf{P}ost-\textbf{T}raining for \textbf{VLA} Model}
\label{alg:main}
\begin{algorithmic}[1]
\Input Pretrained VLA $\pi_\theta$, reward function $R(\mathbf{c}, \mathbf{a})$, context dataset $\mathcal{D}_{\text{context}}$
\For{$\text{step} = 1$ to $M$}
    \State Update sampling VLA $\pi_\psi \leftarrow \pi_\theta$
    \State Initialize empty dataset $\mathcal{D}_{\text{rollout}} \leftarrow \emptyset$
    \While{$|\mathcal{D}_{\text{rollout}}| < B$} \Comment{\textbf{Rollout Collection}}
        \State Sample a context $\mathbf{c} \leftarrow (g, o_1) \sim \mathcal{D}_{\text{context}}$
        \State Generate $K$ rollouts $\{\mathbf{a}_k \sim \pi_\psi(\cdot \mid \mathbf{c})\}_{k=1}^K$  \Comment{\textbf{Group Sampling}}
        \State Compute rewards $\{R_k \leftarrow R(\mathbf{c}, \mathbf{a}_k)\}_{k=1}^K$
            \State Compute baselines: $b_k \leftarrow \frac{1}{K - 1} \sum_{j \ne k} R_j$ \Comment{\textbf{Leave-One-Out Baseline}}
            \State Compute advantages: $A_k \leftarrow R_k - b_k$ for each $k$
        \If{all $A=0$} \Comment{\textbf{Dynamic Rejection}}
            \State \textbf{continue}
        \EndIf
            \State Add $(\mathbf{c}, \mathbf{a}_k, A_k)$ for all $k$ to $\mathcal{D}_{\text{rollout}}$
    \EndWhile
    \For{$\text{iteration} = 1$ to $N$}
        \State Update $\pi_\theta$ with PPO loss (Equation~\ref{equ_ppo}) over  $\mathcal{D}_{\text{rollout}}$ \Comment{\textbf{Policy Optimization}}
    \EndFor
\EndFor
\end{algorithmic}
\end{algorithm}

As mentioned above, there is a gap between the current VLA training paradigm and our essential goal of making it work in our downstream tasks.
On one hand, pure supervised training on offline data makes the policy fragile in real rollout due to compounding errors and the distribution gap between the offline dataset and online rollout.
Furthermore, one has to collect a sufficient number of high-quality demonstrations for offline datasets, especially $\mathcal{D}_{\text{sft}}$, the model can easily overfit to the training distribution.
In other words, optimizing VLA through Equation~\ref{equ:sft} does not necessarily improve the VLA's task execution success rate in Equation~\ref{equ:reward_obj}.
To bridge this gap, we propose a new VLA training paradigm that directly optimize pretrained VLA through interaction with the environment $\mathcal{E}$ through \textbf{R}einforcement \textbf{I}nteractive \textbf{P}ost-\textbf{T}raining, or \METHODNAME for short.

\subsection{Reinforcement Interactive Post-Training for VLA Models}
The first two stages of our VLA training paradigm are the same as the standard setting.
In Stage 1, we pretrain the VLA model on a large, diverse dataset $\mathcal{D}_{\text{pretrain}}$ to learn visual-language representation and general visuomotor skills.
Then, in Stage 2, we finetune VLA on a small dataset $\mathcal{D}_{\text{sft}}$ to adapt it to follow instructions to solve a small set of target tasks. These stages produce a pretrained VLA policy $\pi_\theta$ that can achieve a non-zero success rate (can be very low) on the target tasks.

In \METHODNAME, we then conduct \textbf{Stage 3: Reinforcement Interactive Post-Training}.
In this stage, we assume we can rollout $\pi_\theta$ in an environment $\mathcal{E}$ and receive a binary reward $R(\mathbf{c}, \, \mathbf{a}) \in \{0, 1\}$ given $\mathbf{a} \sim \pi_\theta(\cdot \mid \mathbf{c})$, where $\mathbf{c}$ is the initial context.
In addition, we use an initial context dataset $\mathcal{D}_{\mathbf{c}} = \{(o_1, g)\}$ to set up task initializations for model rollouts.
Typically, we obtain $\mathcal{D}_{\mathbf{c}}$ by directly extracting the initial states from sequences in $\mathcal{D}_{\text{sft}}$.
For each optimization step, we iterate between two steps: \textbf{rollout collection} and \textbf{policy optimization}.

During \textit{rollout collection}, we randomly sample contexts $\mathbf{c}_i \sim \mathcal{D}_{\mathbf{c}}$ and let $\pi_\theta$ interact with the environment $\mathcal{E}$ to output a sequence $\mathbf{a}_i$.
For each rollout we collect its reward $R(\mathbf{c}_i, \mathbf{a}_i)$ and compute its advantage $A_i = A(\mathbf{c}_i, \mathbf{a}_i)$, which indicate how strong the model should be encouraged ($A>0$) or penalized ($A<0$) for generating rollout $\mathbf{a}$.
We add all rollouts and rewards $(\mathbf{c}_i, \mathbf{a}_i, A_i)$ to a rollout dataset $\mathcal{D}_{\text{rollout}}$ until we obtain $B$ rollouts: $\mathcal{D}_{\text{rollout}} = \{(\mathbf{c}_i, \mathbf{a}_i, A_i)\}_{i=1}^B$

During \textit{policy optimization}, we optimize $\pi_\theta$ with reinforcement learning algorithms on $\mathcal{D}_{\text{rollout}}$ to maximize its expected task success rate in Equation~\ref{equ:reward_obj} for $N$ iterations.
After optimization, we use the updated VLA policy $\pi_\theta'$ to collect new rollouts and a new step begins.
This process repeats until we reach $M$ steps and outputs the final policy $\pi_\theta^*$, concluding the full VLA training paradigm.
We then deploy $\pi_\theta^*$ in the environment for testing.

Although \METHODNAME is simple in concept, it presents several challenges.
First, we only have sparse binary rewards from each rollout sequence, no shaped reward is available.
Training a learned reward model to predict shaped reward values can easily lead to reward hacking~\cite{skalse2022defining}, especially with limited rollout data.
Second, as VLA models operate over long-horizon, multi-task environments, credit assignment becomes highly ambiguous.
This causes the value target (e.g., from TD error) to be extremely noisy and uninformative.
Third, training a stable value function for VLA requires a model of comparable capacity to the VLA itself, which significantly increases GPU memory usage and training cost for large VLA models~\cite{zhai2024fine}.
Finally, in multitask environments, different task contexts can vary significantly in difficulty: some lead to trivial success while others consistently fail across all rollouts. This results in highly imbalanced success rates and unstable policy gradient updates.

\subsection{Dynamic-Sampling Leave-One-Out Proximal Policy Optimization}
\label{sec:loop_vla}
To implement \METHODNAME in a stable and sample-efficient way, we propose a simple yet effective policy optimization framework in Algorithm~\ref{alg:main}.
First, we adopt LOOP (Section~\ref{sec_pre_loop}) as the foundation of our implementation.
LOOP is particularly well-suited for our VLA setting, where rollouts are long-horizon and efficient advantage estimation is required for its sparse reward signal.
Furthermore, for VLA in multitask environments, we design a dynamic rollout sampling mechanism to filter out uninformative contexts for more stable and efficient policy optimization.

\mypar{LOOP for \METHODNAME.}
We apply LOOP~\cite{chen2025reinforcementlearninglonghorizoninteractive} for both the rollout collection and policy optimization stage.
During rollout collection, we conduct RLOO~\cite{koolattention} advantage estimation. 
In this step, we use the most recent policy $\pi_\theta$ as the sampling policy $\pi_\psi$.
Given a single context $\mathbf{c} \sim \mathcal{D}_{\mathbf{c}}$, we collect $K$ trajectories by repeatingly sampling $K$ times from the policy given the same context: $\{\mathbf{a}_k \sim \pi_\psi(\cdot \mid \mathbf{c})\}_{k=1}^K$.
We obtain their corresponding rewards $\{R_k\}_{k=1}^K$ from the environment $\mathcal{E}$.
For each rollout $k$, we compute the advantage $A_k$ with Equation~\ref{equ_rloo}.
For each epoch, we conduct group sampling on $B/K$ contexts sampled from $\mathcal{D}_{\mathbf{c}}$, obtaining $\mathcal{D}_{\text{rollout}}$ with $B$ rollouts.

During policy optimization, we use PPO~\cite{schulman2017proximal} to stabilize policy gradient updates.
For each rollout sample $(\mathbf{c}_i, \mathbf{a}_i, A_i) \in \mathcal{D}_{\text{rollout}}$, we can compute its training objective $\mathcal{L}_{\text{PPO}}$ with Equation~\ref{equ_ppo}.
We perform this update over the collected rollout dataset $\mathcal{D}_{\text{rollout}}$ using mini-batches for $N$ optimization steps each epoch. 
When $N = 1$, the method corresponds to on-policy RLOO; when $N > 1$, the same samples are reused for additional updates, resulting in a partially off-policy optimization.

\mypar{Dynamic rollout sampling.} 
VLA models often operate in multitask environments~\cite{kim2024openvla,mete2024quest,sun2022paco}, where task difficulty varies widely across different contexts.
Some contexts have already been well solved by the VLA model, leading to trivial success across $K$-group sampling, while others consistently fail due to inherent task complexity or distribution gap.
Both cases result in rollout groups where all rollout samples receive identical rewards (all ones or all zeros), producing all-zero advantages in Equation~\ref{equ_rloo}.
Therefore there is no gradient signal from Equation~\ref{equ_ppo}.
Adding these samples to $\mathcal{D}_{\text{rollout}}$ makes unstable gradient updates during batch optimization, as they contribute zero gradients that can dilute meaningful learning signals.

To address this, we apply a simple yet effective dynamic rejection strategy: we discard any sampled context for which all $K$ rollouts receive the same reward and resample a new context from $\mathcal{D}_{\text{context}}$ for group sampling.
As training progresses and the policy improves, an increasing number of task contexts yield uniformly successful rollouts. Dynamic rejection naturally filters out these solved contexts, allowing optimization to concentrate on the remaining harder contexts.
Importantly, this method make the batch optimization of the PPO loss (Equation~\ref{equ_ppo}) to have the same effective batch size over all the minibatches across $\mathcal{D}_{\text{rollout}}$, which we empirically found to be important for stable policy optimization in \METHODNAME.

The full implementation of our optimization procedure is summarized in Algorithm~\ref{alg:main}.

\subsection{Generalizing \METHODNAME to Different VLA models}
\label{sec:diff_vlas}
\METHODNAME is compatible with both discrete and continuous action representations commonly used in VLA models.
\METHODNAME requires the VLA model being able to output a probability distribution over actions at each step, which is used in two key steps.
First, during rollout collection, to support diverse rollouts in group sampling for the same initialization context, we need to randomly sample different actions from its output distribution:
\begin{equation}
    a_t \sim \pi_\theta(a_t \mid o_{1:t}, g, a_{1:t-1}),
\end{equation}

On the other hand, to perform stable policy optimization, we compute the trust region $r_i = \frac{\pi_\theta(\mathbf{a}_i \mid \mathbf{c}_i)}{\pi_\psi(\mathbf{a}_i \mid \mathbf{c}_i)}$ in Equation~\ref{equ_ppo} to constrain policy updates within a small region of the original policy.
A key component in this formulation is computing the log-probability of the sampled action sequences under both policies.
We compute the log-probability of a sampled action sequence $\mathbf{a} = (a_1, \dots, a_T)$ as the sum of the per-step log-probabilities:
\begin{equation}
\log \pi_\theta(\mathbf{a} \mid \mathbf{c}) = \sum_{t=1}^T \log \pi_\theta(a_t \mid a_{<t}, \mathbf{c}).
\end{equation}

In other words, we can apply \METHODNAME to any VLA model $\pi_\theta$ that we sample a random action $a_t$ from the per-step action distribution and compute $\log \pi_\theta(a_t \mid a_{<t}, \mathbf{c})$.

\mypar{Tokenized action head.}
For VLA models with discrete action outputs,~\eg QueST~\cite{mete2024quest},
actions are predicted as sequences of discrete tokens from a fixed vocabulary, where the action header is a classification head trained with NLL loss.
Therefore, $\log \pi_\theta(a_t \mid a_{<t}, \mathbf{c})$ is directly obtained from applying softmax function to the model’s classification head output logits.
We can also simply sample action tokens from the distribution after softmax.

\mypar{Regression action head.}
For continuous-action VLA models~\cite{kim2025openvlaoft}, actions are regressed using MSE or L1 loss, which do not produce a log-probability.
To enable policy gradient optimization, we extend the model with a light-scale prediction head that estimates the scale $\sigma_\theta$ of the action value.
Assuming the original output head provides the mean $\mu_\theta$, we treat the policy as a factorized Gaussian (MSE) or Laplace (L1) distribution and train the scale head using the NLL loss in Equation~\ref{equ:sft} for a few iterations on $\mathcal{D}_{\text{sft}}$.
After that, we can sample action $a_t$ and compute $\log \pi_\theta(a_t \mid a_{<t}, \mathbf{c})$ with predicted $\mu_\theta$ and $\sigma_\theta$ in a closed form.

\section{Experiments}
\label{sec:exp}
We evaluate \METHODNAME on two widely used benchmarks for VLA learning: LIBERO~\cite{liu2023libero} and MetaWorld~\cite{yu2020meta}.
We study several settings: (1) standard multitask (up to 90 tasks) setting in Sec.~\ref{exp:standard}, (2) few-shot ($1\sim5$ demonstration) setting in Sec.~\ref{exp:fewshot}, and (3) cross-task and cross-scenario setting in Secs.~\ref{exp:cross_scene} and~\ref{exp:cross_goal} to showcase the ability of fast generalization leveraging prior knowledge during pretraining.
Additionally, we conducted studies to analyze the practical behavior of \METHODNAME, including training curves, ablation studies as well as its sensitivity to the variance and diversity of the context dataset.

\subsection{Setup}
\mypar{Benchmark.}
LIBERO~\cite{liu2023libero} is a lifelong learning benchmark with 5 task suites.
Each suite consists of a set of language-guided manipulation tasks across multiple object types, task definitions and environment scenarios.
Specifically, it includes 4 suites: \textbf{Goal}, \textbf{Spatial}, \textbf{Object}, and \textbf{Long}. Each suite is designed to evaluate a specific aspect of object manipulation and containing 10 distinct tasks.
In addition, it also includes a \textbf{LIBERO-90} suite that contains 90 different tasks to assess multitask performance at scale.
MetaWorld~\cite{yu2020meta} is a manipulation task benchmark for few-shot learning models.
We use Meta-Learning 45 (ML45) suite that contains 45 training tasks and 5 held-out tasks.

For both benchmarks, each task comes with 50 expert demonstrations for training.
At evaluation time, a single VLA model is deployed across all tasks in a suite and performs rollouts on 50 held-out test contexts per task. We measure performance with the average task success rate (SR).

\mypar{Base models.}
We conduct \METHODNAME on two pretrained VLA models with different design choices.

OpenVLA-OFT~\cite{kim2025openvlaoft} is an \textit{Optimized Fine-Tuned} variant of the 7B OpenVLA model~\cite{kim2024openvla}.  
OpenVLA is initialized from a multimodal backbone that combines a \textit{Llama-2 7B} language model with dual vision encoders~\cite{oquab2023dinov2,zhai2023sigmoid} 
and is pretrained on 970k robot-manipulation demonstrations.
OFT replaces the original tokenized action decoder with a continuous decoding head and trains with an L1 regression loss.
This architecture represents the \textit{large-scale regression action} VLA.

QueST~\cite{mete2024quest}, on the other hand, is a \textit{small-scale tokenized action} VLA model with 20 million parameters.
QueST first learns a VQ-VAE that compresses short motion segments into a discrete \textit{skill codebook}; a GPT-style transformer then autoregressively predicts these skill tokens conditioned on images and language, and a small decoder turns tokens back into continuous joint commands.  

\mypar{Implementation details.}
For OpenVLA-OFT, we fine-tune the model using the official checkpoints provided for each task suite. Training is conducted on 4 NVIDIA RTX A5000 GPUs, each with 24 GB memory.
We use LoRA~\cite{hu2022lora} with rank 32, and set $K=8$, $B=192 (8\times24)$, $N=1$ and $\epsilon=0.1$, PPO mini-batch size of 4 per GPU.
We set a learning rate of $1\mathrm{e}{-4}$ for the LoRA modules and $1\mathrm{e}{-5}$ for the action head.
Following Section~\ref{sec:diff_vlas}, before applying \METHODNAME, we first train a small Laplace scale header from scratch (with the same architecture as the action header) with NLL loss on $\mathcal{D}_{\text{sft}}$ for 500 steps.
For rollout collection during training, we randomly sample actions according to the Laplace distribution with scale predicted by this header.
For evaluation, we directly use the mean value predicted by the original action header.

For QueST, as official checkpoints are not provided, we first train the base model from scratch for each task suite following the official code and hyperparameters.
In the multitask setting, we conduct \METHODNAME on 3 GPUs with $K=16$, $B=2880$ ($16\times180$).
For the single-task setting, we use 1 GPU with $K=16$, $B=160$.
For both settings, we set $N=20$, PPO mini-batch size of 8 per GPU, a learning rate of $1\mathrm{e}{-6}$, and the clipping parameter $\epsilon=0.2$.

\subsection{Standard Multitask Training}
\label{exp:standard}

\begin{table*}[t]
\centering
\small
\begin{tabular}{lccccc}
\toprule
\multicolumn{6}{c}{\textbf{Stage 1 + Stage 2 Models}} \\
\midrule
\textbf{Method} & \textbf{Goal} & \textbf{Spatial} & \textbf{Object} & \textbf{Long} & \textbf{Average} \\
\midrule
Octo~\cite{team2024octo} & 84.6 & 78.9 & 85.7 & 51.1 & 75.1 \\
OpenVLA~\cite{kim2024openvla} & 79.2 & 84.7 & 88.4 & 53.7 & 76.5 \\
Dita~\cite{hou2025dita} & 85.4 & 84.2 & 96.3 & 63.8 & 82.4 \\
$\pi_0$ + FAST~\cite{pertsch2025fast} & 88.6 & 96.4 & 96.8 & 60.2 & 85.5 \\
$\pi_0$~\cite{black2024pi0} & 95.8 & 96.8 & \textbf{98.8} & 85.2 & 94.2 \\
OpenVLA-OFT*~\cite{kim2025openvlaoft} & \underline{97.9} & \underline{97.6} & 98.4 & \underline{92.9} & \underline{96.7} \\
\textbf{OpenVLA-OFT + \METHODNAMESHORT} & \textbf{99.0} \textbf{\textcolor{myred}{(+1.1)}} & \textbf{98.6} \textbf{\textcolor{myred}{(+1.0)}} & \underline{98.6} \textbf{\textcolor{myred}{(+0.2)}}& \textbf{93.8} \textbf{\textcolor{myred}{(+0.9)}} & \textbf{97.5} \textbf{\textcolor{myred}{(+0.8)}} \\
\midrule
\multicolumn{6}{c}{\textbf{Stage-2 Models}} \\
\midrule
\textbf{Method} & \textbf{Goal} & \textbf{Spatial} & \textbf{Object} & \textbf{Long} & \textbf{Average} \\
\midrule
Diffusion Policy~\cite{chi2023diffusion} & 68.3 & 78.3 & 92.5 & 50.5 & 72.4 \\
Seer~\cite{tian2024predictive} & -- & -- & -- & 78.7 & -- \\
MDT~\cite{reuss2024multimodal} & 73.5 & 78.5 & 87.5 & 64.8 & 76.1 \\
MDT+~\cite{reuss2024multimodal} & -- & \underline{95.2} & \underline{97.8} & \underline{83.0} & -- \\
QueST~\cite{mete2024quest} & \underline{80.8} & 87.4 & 93.6 & 68.8 & 82.7 \\
\textbf{QueST + \METHODNAMESHORT} & \textbf{92.7} \textbf{\textcolor{myred}{(+11.9)}} & \textbf{95.6} \textbf{\textcolor{myred}{(+8.2)}} & \textbf{98.4} \textbf{\textcolor{myred}{(+4.8)}} & \textbf{87.5} \textbf{\textcolor{myred}{(+18.7)}} & \textbf{93.6} \textbf{\textcolor{myred}{(+10.9)}} \\
\bottomrule
\end{tabular}
\caption{Multitask SR(\%) on the four LIBERO suites. 
\textbf{Bold} indicates best result and \underline{underline} marks the second-best.
Improvements from \METHODNAME are \textcolor{myred}{\textbf{marked in red}}. *: OpenVLA-OFT results are obtained from running official checkpoints for each suite.}
\label{tab:libero-four-tasks}
\end{table*}

In this section, we evaluate \METHODNAME under standard multitask benchmarks.
For each suite, we use all 50 expert demonstrations per task as its SFT dataset $\mathcal{D}_{\text{sft}}$.
We conduct \METHODNAME to finetune a base model on the corresponding dataset for each task suite.

Table~\ref{tab:libero-four-tasks} compares multitask performance on four LIBERO suites for different VLA models.
We organize the results into two sets based on VLA training paradigm.
In the \textbf{Stage 1+ Stage 2} set, we include 5 state-of-the-art large VLA models: Octo~\cite{team2024octo}, OpenVLA~\cite{kim2024openvla}, Dita~\cite{hou2025dita}, $\pi_0$~\cite{black2024pi0} and OpenVLA-OFT~\cite{kim2025openvlaoft}.
These models are typically larger than 500M parameters, pretrained (Stage-1) on large-scale general-purpose datasets, \textit{e.g.,} Open-X Embodiment~\cite{o2024open}, and then finetuned using 50 demonstrations per task for each LIBERO suite (Stage-2).
In contrast, the \textbf{Stage 2} set includes 4 representative small models: Diffusion Policy~\cite{chi2023diffusion}, Seer~\cite{tian2024predictive}, MDT~\cite{reuss2024multimodal} and QueST~\cite{mete2024quest}.
These models are within 50M parameters and are directly trained on each LIBERO suite from scratch.

We show that \METHODNAME significantly improves the best-performing VLA model in both types, setting new state-of-the-art performance on the 4 LIBERO suites.
Specifically, \METHODNAME improves QueST on all four task suites by \textbf{10.9} absolute SR on average, and yields even larger gains of \textbf{18.7} for the challenging LONG suite.
Notably, with \METHODNAME, the small 20M QueST model achieves much better performance with large models like Dita (334M) and comparable with $\pi_0$ (2B).
When applying to OpenVLA-OFT, the best-performing large VLA model with already high SR, \METHODNAME still further reduces the average failure rate from 3.3\% to 2.4\%.
By applying \METHODNAME, we set new state-of-the-art performance on 3 out of the 4 LIBERO suites (with only a 0.2 gap on the Object suite), and achieve the highest average success rate across all tasks.
These results show the \METHODNAME is broadly effective: it can both unlock latent capabilities in small-scale models and further push the limits of the high-performing ones.

In addition, in the left two columns of Table~\ref{tab:all-benchmark-results}, we show the results on LIBERO-90 and ML45, which contain 90 and 45 diverse tasks respectively.
These benchmarks assess the scalability and generalization of a single VLA model across many skills.
We apply \METHODNAME to QueST and compare with representative imitation learning methods: ACT~\cite{gao2024act}, PRISE~\cite{zheng2024prise}, Diffusion Policy~\cite{chi2023diffusion}, VQ-BeT~\cite{lee2024behavior} and ResNet-T~\cite{mete2024quest}.
We show that \METHODNAME improves performance of QueST by \textbf{5.7} and \textbf{1.2} absolute SR for LIBERO-90 and ML45, again setting new SOTA performance for both benchmarks.
This confirms the utility of \METHODNAME not only for improving performance on a few related tasks, but also for scaling up to broader, more realistic scenarios where a single model solves many different tasks.

\begin{table*}[t]
\centering
\begin{minipage}{0.68\textwidth}
    \small
    \begin{tabular}{lcccc}
    \toprule
     & \multicolumn{2}{c}{\textbf{Full Data}} & \multicolumn{2}{c}{\textbf{5-shot Data}} \\
    \textbf{Method} & \textbf{LIBERO-90} & \textbf{ML45} & \textbf{LONG} & \textbf{ML45} \\
    \midrule
    ACT~\cite{gao2024act} & 50.8 & 90.8 & 42.0 & \underline{70.8} \\
    PRISE~\cite{zheng2024prise} & 54.4 & 80.4 & \underline{52.7} & 66.8 \\
    DP~\cite{chi2023diffusion} & 75.4 & 90.3 & 45.9 & 65.0 \\
    VQ-BeT~\cite{lee2024behavior} & 81.3 & 87.6 & 41.8 & 65.6 \\
    ResNet-T~\cite{mete2024quest} & 84.4 & 88.4 & 51.9 & 54.0 \\
    QueST~\cite{mete2024quest} & \underline{88.6} & \underline{91.0} & 50.2 & 63.6 \\
    \textbf{QueST + \METHODNAMESHORT} & \textbf{94.3} & \textbf{92.2} & \textbf{71.4} & \textbf{76.0} \\
    (improvement) & \textbf{\textcolor{myred}{(+5.7)}} & \textbf{\textcolor{myred}{(+1.2)}} & \textbf{\textcolor{myred}{(+21.2)}} & \textbf{\textcolor{myred}{(+12.4)}} \\
    \bottomrule
    \end{tabular}
    \captionsetup{justification=raggedright, singlelinecheck=false, margin=-1em}
    \caption{
        Mean Success Rate (SR\%) across four evaluation settings:
        LIBERO-90 and ML45 (Full data), LONG and ML45 (5-shot).
    }
    \label{tab:all-benchmark-results}
\end{minipage}
\hfill
\begin{minipage}{0.3\textwidth}
    \vspace{1mm}
    \hspace*{-2.3em}\includegraphics[height=3.9cm]{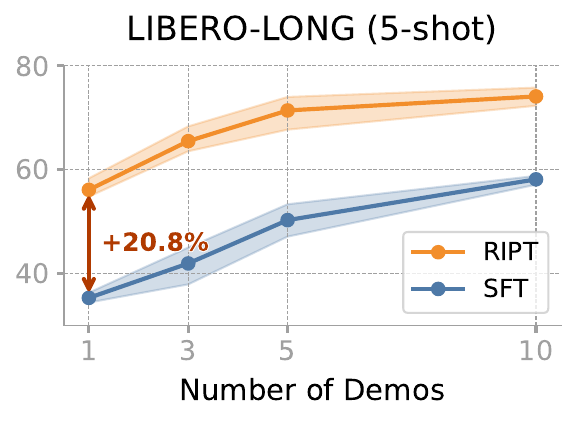}
    \vspace{-4.3mm}
    \captionsetup{justification=raggedright, singlelinecheck=false, margin=1em}
    \captionof{figure}{Few-shot curve on LIBERO-LONG.}
    \label{fig:fewshot}
\end{minipage}
\end{table*}

\subsection{Few-shot Multitask Training}
\label{exp:fewshot}

In this section we evaluate \METHODNAME under few-shot multitask setting.
For each suite, we uniformly sample 1 to 10 expert demonstrations from each task to constitute the few-shot SFT dataset $\mathcal{D}_{\text{sft}}$.
This setting reflects practical situation where large-scale data collection is not available.

The right two columns of Table~\ref{tab:all-benchmark-results} show results under the 5-shot setting, where each task in the LIBERO-LONG and ML45 suites is trained with only 5 demonstrations.
While baseline models struggle in this low-data regime, \METHODNAME significantly improves QueST by \textbf{21.2} on LIBERO-LONG and \textbf{12.4} on ML45.
These results demonstrate that \METHODNAME effectively addresses a key limitation of standard VLA training with SFT: it enables strong performance even with minimal demonstrations, alleviating concerns about data scarcity in real-world multitask deployment.

To further investigate the effect of the number of few-shot demonstrations, we conduct experiments under varying few-shot settings with QueST, ranging from 1 to 10 demonstrations per task on LIBERO-LONG.
As shown in Figure~\ref{fig:fewshot}, \METHODNAME consistently largely improves the performance of the standard SFT model across all data scales.
Note that even for the extremely low-data regime, where we only have 1 demonstration per task, \METHODNAME can still achieve a \textbf{20.8} absolute gain.
As the number of demonstrations increases, \METHODNAME continues to yield performance improvements, indicating its strong sample efficiency and scalability.
These results
confirm that \METHODNAME is robust across different levels of data scarcity and is applicable in both low- and high-data settings.

\subsection{Cross-scenario Generalization}
\label{exp:cross_scene}

\begin{figure}[t]
    \centering
    \includegraphics[width=1.0\textwidth]{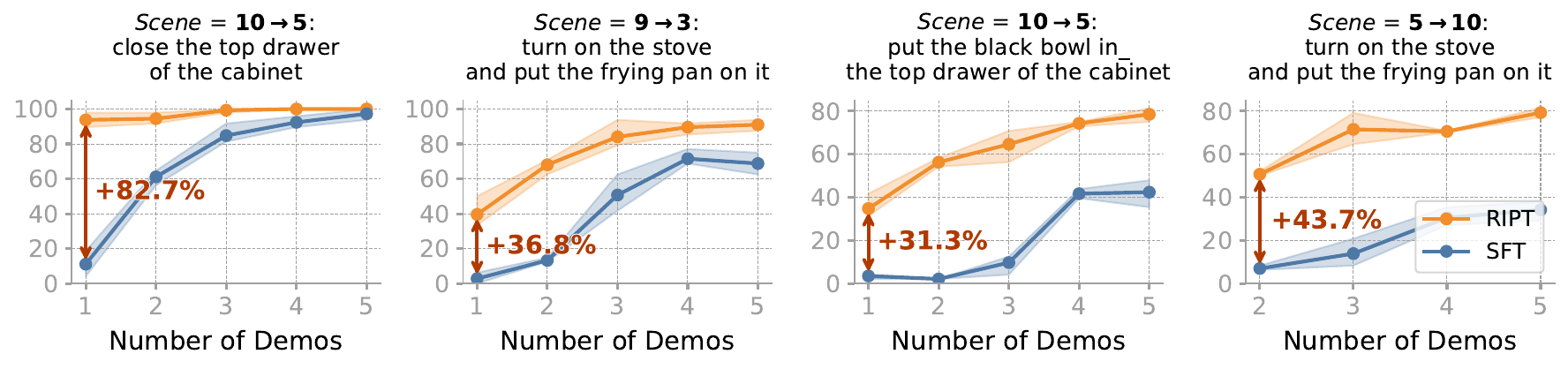}
    \caption{Cross-scenario task generalization from Scenario A to Scenario B with the same goal.}
    \label{fig:cross_scenario_all}
\end{figure}

\begin{figure}[t]
    \centering
    \includegraphics[width=1.0\textwidth]
    {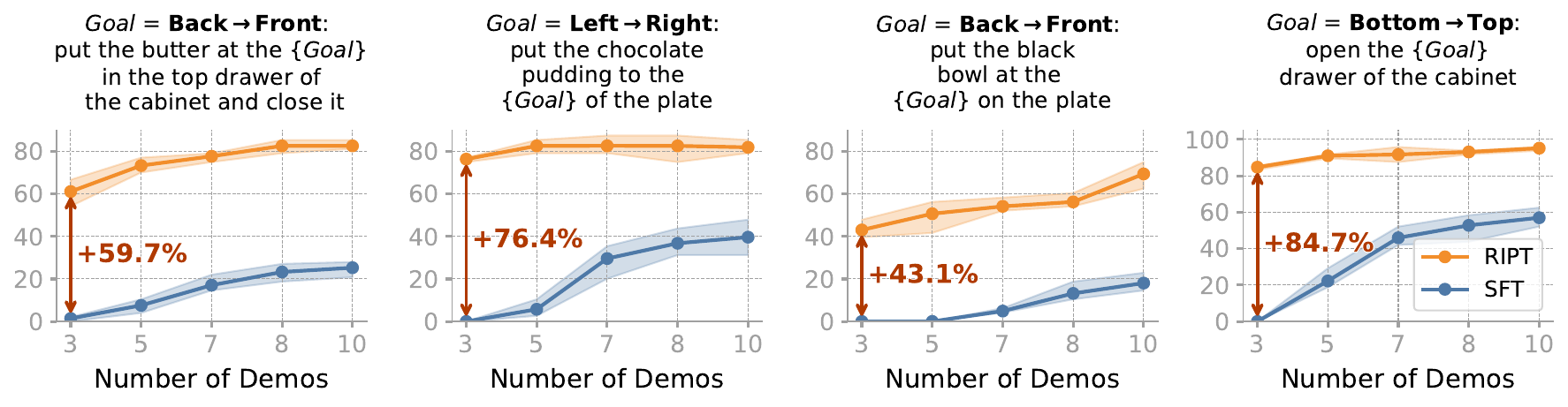}
    \caption{Cross-goal task generalization from Goal A to Goal B in the same scenario.}
    \label{fig:cross_goal_all}
\end{figure}

Recent paradigm shift in LLM training demonstrates that reinforcement learning can reactivate and steer pretrained knowledge with only a small amount of interactive feedback~\cite{ouyang2022training}.
We adopt a similar approach for VLA and ask: can \METHODNAME enable sample-efficient pretrained visuomotor skill transfer across scenarios and goals?

In this section, we experiment on the few-shot cross-scenario generalization setup.
For each experiment, we consider a pair of tasks that have the same task goal (e.g., \textit{'turn on the stove and put the frying pan on it'}), but operate in different scenarios: Scenario A and Scenario B - with distanct background layouts and object configurations.
In Stage 1, we pretrain QueST on $|\mathcal{D}_{\text{pretrain}}|=50$ demonstrations from Scenario A to acquire a general visuomotor skill for this task goal.
In Stage 2, we conduct SFT on $|\mathcal{D}_{\text{sft}}|=\{1,2,3,4,5\}$ demonstrations from Scenario B.
Then, in Stage 3, we apply \METHODNAME to optimize the policy through interactive rollouts on contexts $\mathcal{D}_{\text{context}}$ extracted from $\mathcal{D}_{\text{sft}}$.
We then evaluate the model on the 50 testing contexts of Scenario B.
We conduct experiments with 3 random seeds and plot the mean and variance across different sizes of $\mathcal{D}_{\text{sft}}$.

Figure~\ref{fig:cross_scenario_all}
show results on 5 scenario pairs.
We observe that standard SFT on VLA models clearly struggles in the \textbf{1-shot} regime, achieving an average success rate of only around 5\%, and in some cases dropping as low as 2\%.
Clearly, SFT fails to generalize the task knowledge from the pretraining stage to the new scenario.
In contrast, \METHODNAME dramatically improves performance, with absolute SR gain as high as \textbf{93.7\%} (from 3.5\% SFT to 97.2\%).
As the size of $\mathcal{D}_{\text{sft}}$ increases, both SFT and \METHODNAME performance improve, but \METHODNAME consistently maintains a strong improvement, often reaching near-100\% performance with just 3-5 demonstrations.
These results supports our core assumption: \METHODNAME enables pretrained VLA models to rapidly activate and adapt learned skills with sparse binary reward feedback.

\subsection{Cross-goal Generalization}
\label{exp:cross_goal}
In this section, we investigate \METHODNAME in a cross-goal generalization setting.
Here we focus on task pairs that operate in the same scenario but with different goals.
Specifically, we select Task A and Task B such that they require the same visuomotor skills but have different goals.
For example, Task A is \textit{"put the red mug on the \textbf{right} plate"} while Task B is \textit{"put the red mug on the \textbf{left} plate"}.
This setting tests whether pretrained visuomotor primitive skills (e.g., pick up and move) can be reused and recomposed to solve novel task goals (e.g., left vs. right).
We again follow the 3-Stage paradigm: pretrain QueST on 50 demonstrations of Task A, SFT on 3-10 demonstrations on Task B, and then apply \METHODNAME for Task B.

Figure~\ref{fig:cross_goal_all} presents results over 5 sets of tasks.
We observe that cross-goal generalization is significantly more challenging.
With 3 demonstrations, SFT models still struggle and reach only \textbf{0.7\%} success rate on average, almost not workable at all.
With \METHODNAME, we can improve model performance to \textbf{59.7\%} on average.
Remarkably, for one task pair, \METHODNAME improves the performance from near \textbf{0\%} success rate to \textbf{84.7\%}.
As the number of demonstrations increases, \METHODNAME consistently maintains a large advantage across all data regions.
At 10 demonstrations, the average success rate of \METHODNAME reaches \textbf{79.7\%}, compared to only \textbf{29.4\%} for SFT.

\begin{table*}[t]
\centering
\begin{minipage}{0.33\textwidth}
    \centering
    \includegraphics[width=1.0\textwidth]{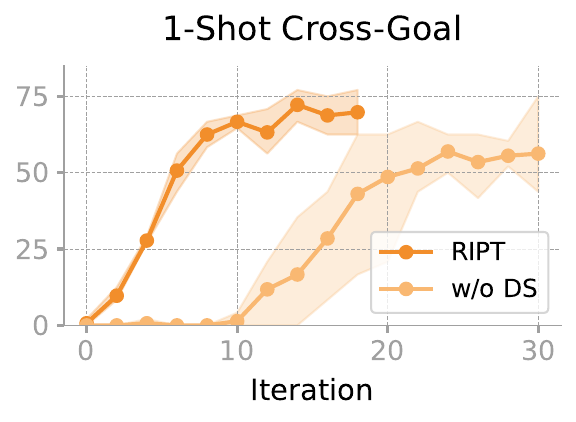}
    \vspace{-5.5mm}
    \captionsetup{justification=raggedright, singlelinecheck=false, margin=0em}
    \captionof{figure}{Training curve analysis of dynamic sampling.}
    \label{fig:train_curve}
\end{minipage}
\hfill
\begin{minipage}{0.32\textwidth}
    \centering
    \includegraphics[width=1.0\textwidth]{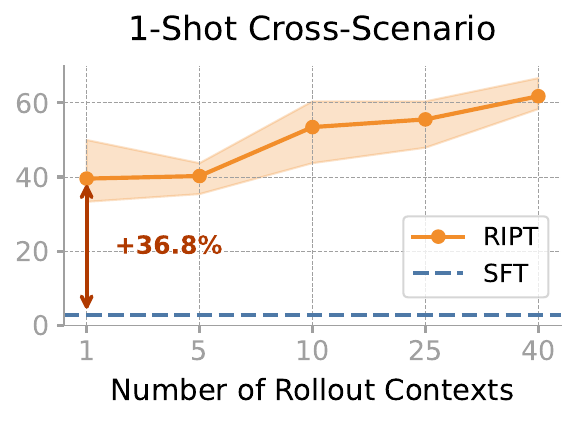}
    \vspace{-5.5mm}
    \captionsetup{justification=raggedright, singlelinecheck=false, margin=0em}
    \captionof{figure}{Analysis on context dataset size.}
    \label{fig:context_dataset_size}
\end{minipage}
\hfill
\begin{minipage}{0.32\textwidth}
    \centering
    \includegraphics[width=1.0\textwidth]{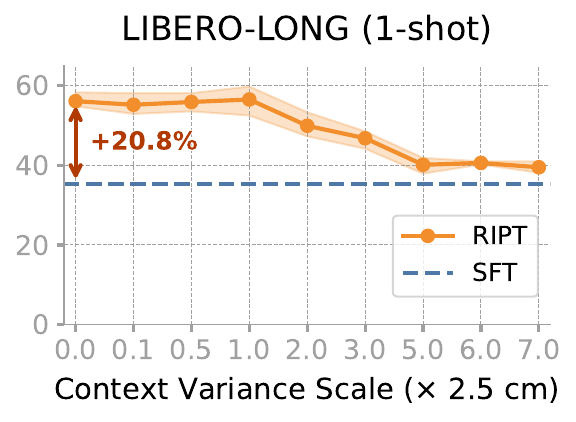}
    \vspace{-5.5mm}
    \captionsetup{justification=raggedright, singlelinecheck=false, margin=0em}
    \captionof{figure}{Analysis on initial state std scale}
    \label{fig:variance_scale}
\end{minipage}
\end{table*}

These results further show the limitation of SFT paradigm for VLA generalization under low-data regime.
In contrast, we show that \METHODNAME not only helps adapt pre-trained skills to new environments, but also excels in fast generalization of task goal semantics.

\begin{table*}[t]
\centering
\small
\begin{tabular}{lccccccc}
\toprule
\textbf{Method} & \textbf{Goal} & \textbf{Spatial} & \textbf{Object} & \textbf{Long} & \textbf{90} & \textbf{ML45} & \textbf{Average} \\
\midrule
QueST & 80.8 & 87.4 & 93.6 & 68.8 & 88.6 & 91.0 & 85.0 \\
+ \METHODNAME w/o Dynamic Sampling & 90.6 & 91.3 & 97.5 & 78.3 & 92.2 & 91.3 & 90.2 \\
\textbf{+ \METHODNAME (Ours)} & \textbf{92.7} & \textbf{95.6} & \textbf{98.4} & \textbf{87.5} & \textbf{94.3} & \textbf{92.2} & \textbf{93.5} \\
\bottomrule
\end{tabular}
\caption{Ablation on dynamic sampling. We compare full \METHODNAME against a variant without dynamic sampling and the QueST baseline across task types and multitask suites.}
\label{tab:ablation-dynamic-sampling}
\end{table*}

\subsection{Aditional Study}


\mypar{Effect of dynamic rollout sampling.}
We ablate the impact of our dynamic rollout sampling strategy described in Section~\ref{sec:loop_vla}.
We compare the full \METHODNAME method with a variant that disables dynamic rejection. 
As shown in Table~\ref{tab:ablation-dynamic-sampling}, dynamic sampling significantly boosts performance across all task categories and suites.
By filtering out uninformative rollout groups, dynamic sampling ensures stable and efficient learning with a consistent gradient signal across batches. 
On average, we observe a \textbf{+3.3} absolute improvement in success rate compared to the non-dynamic variant, demonstrating its crucial role in stabilizing \METHODNAME training.
In Figure~\ref{fig:train_curve}, we show the training curve (averaged over 3 seeds) of Column 2 of Figure~\ref{fig:cross_goal_all}.
We see that dynamic rollout sampling accelerates convergence of \METHODNAME, achieving consistently higher performance and more stable optimization.

\mypar{Effect of context dataset size.}
To study how the size of the context dataset $\mathcal{D}_{\mathbf{c}}$ impacts performance, we fix the QueST model SFT-trained with only 1 demonstration for Column 2 of Figure~\ref{fig:cross_scenario_all} and vary the number of rollout contexts used in the \METHODNAME stage.
As shown in Figure~\ref{fig:context_dataset_size}, increasing the number of rollout contexts significantly improves performance. 
This is because more contexts provide greater diversity in initial states for the rollout interaction, allowing the model to better generalize across different setups in the testing environments.
Notably, expanding $\mathcal{D}_{\mathbf{c}}$ requires no additional human annotations: each context only consists of the initial observation state and no action is needed.
This makes context dataset scaling a cost-effective way to enhance generalization of \METHODNAME.

\mypar{Effect of context variance in RLOO group.}
In Equation~\ref{equ_rloo}, each batch of rollouts is grouped by shared initial state contexts.
In realistic deployments, however, perfectly matching initial states is impractical due to inevitable setup noise. 
To simulate this, we compute the standard deviation of object initial positions across LIBERO-LONG, which is around 2.5 cm.
Starting with a QueST model SFT on 1 demo, we run \METHODNAME while injecting Gaussian noise into the initial states with increasing scales of std.
As shown in Figure~\ref{fig:variance_scale}, performance remains stable up to 1.0× (2.5 cm), and only begins to degrade beyond 2.0×. Remarkably, even at 7.0× variance (17.5 cm), \METHODNAME still outperforms the SFT baseline by a significant margin.

\section{Conclusion}
We presented \textbf{\METHODNAME}, a simple yet powerful reinforcement learning paradigm for post-training pretrained Vision-Language-Action (VLA) models using sparse binary task rewards. 
\METHODNAME enables stable and data-efficient optimization without the need for shaped rewards, value functions, or reward modeling. 
Our method significantly improves performance across multiple VLA benchmarks and demonstrates remarkable adaptability even in extremely low-data settings. 
\METHODNAME serves as a scalable third-stage training paradigm that complements existing pretraining and supervised fine-tuning pipelines, unlocking the latent potential of large VLA models through direct environment interaction.
An exciting future direction is to combine \METHODNAME with reasoning and planning in VLA models to enable more sophisticated and generalizable behaviors in complex environments.

\bibliographystyle{plain}
\bibliography{reference}

\end{document}